\documentclass[authorversion,sigconf]{acmart}

\usepackage{placeins}
\usepackage{makecell}
\usepackage{multirow}
\usepackage{siunitx}
\usepackage{dirtree}
\usepackage{subcaption}

\usepackage{stfloats}

\usepackage{fancyhdr}
\AtBeginDocument{%
    \addtolength{\footskip}{2.0\baselineskip}%
    \fancyfoot[L]{\textit{\textbf{Preprint --- Author Accepted Version.}}}%
}

\copyrightyear{2026}
\acmYear{2026}
\setcopyright{cc}
\setcctype{by}
\acmConference[MMSys '26]{ACM Multimedia Systems Conference 2026}{April 04--08, 2026}{Hong Kong, Hong Kong}
\acmBooktitle{ACM Multimedia Systems Conference 2026 (MMSys '26), April 04--08, 2026, Hong Kong, Hong Kong}
\acmDOI{10.1145/3793853.3799806}
\acmISBN{979-8-4007-2481-7/2026/04}

\begin{document}

\title{MobileMold: A Smartphone-Based Microscopy Dataset for Food Mold Detection}

\author{Dinh Nam Pham}
\authornotemark[1]
\orcid{0000-0002-9431-5614}
\affiliation{%
  \institution{Technical University of Berlin}
  \city{Berlin}
  \country{Germany}
}
\email{dinh-nam.pham@campus.tu-berlin.de}

\author{Leonard Prokisch}
\authornotemark[1]
\orcid{0009-0001-8415-6218}
\affiliation{%
  \institution{University of Regensburg}
  \city{Regensburg}
  \country{Germany}}
\email{leonard.prokisch@stud.uni-regensburg.de}

\author{Bennet Meyer}
\authornotemark[1]
\orcid{0009-0001-3415-3588}
\affiliation{%
  \institution{ETH Zurich}
  \city{Zurich}
  \country{Switzerland}}
\email{benmeyer@student.ethz.ch}

\author{Jonas Thumbs}
\orcid{0009-0006-7135-6858}
\affiliation{%
  \institution{University of Tübingen}
  \city{Tübingen}
  \country{Germany}}
\email{jonas.thumbs@student.uni-tuebingen.de}

\authornote{Authors contributed equally to this research.}

\renewcommand{\shortauthors}{Pham, Prokisch, Meyer, Thumbs}

\begin{abstract}
Smartphone clip-on microscopes turn everyday devices into low-cost, portable imaging systems that can even reveal fungal structures at the microscopic level, enabling mold inspection beyond unaided visual checks. In this paper, we introduce MobileMold, an open smartphone-based microscopy dataset for food mold detection and food classification. MobileMold contains 4,941 handheld microscopy images spanning 11 food types, 4 smartphones, 3 microscopes, and diverse real-world conditions.
Beyond the dataset release, we establish baselines for (i) mold detection and (ii) food-type classification, including a multi-task setting that predicts both attributes. Across multiple pretrained deep learning architectures and augmentation strategies, we obtain near-ceiling performance (accuracy = 0.9954, F1 = 0.9954, MCC = 0.9907), validating the utility of our dataset for detecting food spoilage.
To increase transparency, we complement our evaluation with saliency-based visual explanations highlighting mold regions associated with the model’s predictions.
MobileMold aims to contribute to research on accessible food-safety sensing, mobile imaging, and exploring the potential of smartphones enhanced with attachments.
\end{abstract}




\maketitle

\section{Introduction}

Nowadays, smartphones have become ubiquitous in daily life, providing a diverse multitude of multimedia functionalities to the end-user. Consequently, these devices represent a valuable mobile source of high-fidelity data, as high-resolution images, audio, and videos can be captured in almost any environment. However, despite the continuous improvement of mobile hardware, smartphone cameras still have certain constraints, such as limited magnification. To address these limitations, various gadgets have been developed recently to augment the capabilities of mobile phones. Among these are smartphone microscope attachments, which are low-cost lenses that are clipped over the camera to serve as a magnifier. These attachments facilitate mobile microscopy, enabling its application across a wide range of scientific and consumer domains \cite{Raju2024-au}.

For instance, we see great promise for smartphone-based microscopy in food safety applications, particularly for detecting food mold. Conventional methods for detecting fungal contamination rely on culturing microorganisms, which is time-consuming (requiring several days for presumptive results), labor-intensive, and consumable-heavy \cite{Soni2022-bq}. In contrast, while non-destructive approaches such as hyperspectral imaging have demonstrated success \cite{Soni2022-bq}, they require specialized sensors that are relatively expensive and inaccessible for the average consumer. Therefore, we propose smartphone-based microscopy as a non-destructive, low-cost, and rapid alternative for food mold detection. These clip-on lenses are commercially available at low prices, and the image-capturing process requires no expert knowledge. This accessibility allows for deployment in remote locations and day-to-day scenarios, further democratizing the field of imaging.

To this end, we introduce MobileMold, a comprehensive dataset consisting of smartphone-based microscopic images across multiple food categories, including both healthy and contaminated samples. We validate the utility of our dataset by performing food type classification and mold detection using image classification models. Furthermore, we integrate these models into a smartphone application, demonstrating a real-world use case that facilitates mobile food safety assessment for consumers. Because smartphone microscopes significantly enhance the capabilities of mobile cameras, they hold immense potential for various multimedia applications. As a rare resource of smartphone-based microscopic data, MobileMold serves as a valuable asset not only for food safety research but also for the broader multimedia community. The dataset, pre-processing scripts, and code of the experiments are publicly available at \url{https://mobilemold.github.io/dataset/}.

\section{Related Works}

The integration of computer vision and mobile sensing has significantly contributed to the field of food safety and quality assessment. Recent advancements in deep learning have enabled highly accurate identification of fungal contamination on food surfaces using non-destructive image data. Jubayer et al. \cite{Jubayer2021-od} demonstrated the usage of the YOLOv5 architecture for detecting mold on various food surfaces, including fruits and bakery products. By training on a dataset of 2,050 macroscopic images, their model achieved a recall of 100\% and a mean average precision of 99.60\%. However, this approach relies on standard, non-magnified images, which limits its utility to cases where mold growth is already visible to the naked eye. Our work, MobileMold, seeks to address this gap through microscopic analysis.

The ubiquity of smartphones has facilitated the development of "on-site" food monitoring tools that democratize food safety for non-expert consumers. In \cite{Dogan2024-tw}, the authors developed a real-time spoilage monitoring system for fish that utilizes a smartphone-embedded machine learning classifier. Their system employs colorimetric sensing films made from anthocyanin-rich red cabbage extract that react to byproducts of microbial spoilage. By embedding a Random Forest model into a custom Android application, they achieved 99.6\% accuracy in detecting fish freshness. As this relies on chemical spoilage indicators, our work represents a mobile sensing approach facilitating direct visual identification of fungal structures using microscopic attachments.

To detect contamination before it becomes visible to the human eye, researchers have turned to low-cost optical extensions for mobile devices. Treepong and Theera-Ampornpunt \cite{treepongEarlyBreadMold2023b} proposed a method for the early detection of bread mold using $50\times$ magnification clip-on lenses paired with convolutional neural networks. By utilizing transfer learning with architectures like Inception and ResNet, they achieved F1-scores as high as 0.9972 for whole wheat bread. Their study highlights that while professional microscopes are too cumbersome for consumer use, clip-on lenses provide a viable, low-cost alternative for nonprofessionals. Thus, their study is the most conceptually similar to MobileMold. While their work focused specifically on bread, our dataset expands this scope to multiple food types and smartphone microscopes, providing a more diverse and comprehensive resource for the multimedia community.

\section{Dataset Creation}

In this section, we describe our methodology of obtaining the data and provide a description of the MobileMold dataset.

\subsection{Equipment and Setup}

\subsubsection{Microscopy Hardware and Device Pairing}
We utilized commercial smartphone microscope attachments with two different nominal magnifications:

\pagebreak
\begin{enumerate}
    \item $30\times$ nominal magnification: The $30\times$ microscope by Jiusion \cite{Jiusion30XZoom} was paired with a Xiaomi Redmi Note 13 (rear cameras). Additionally, a $30\times$ microscope by Phonescope \cite{led-grower.euPhonescope30xZoom} was paired with a Samsung Galaxy S8+ (main rear camera) and a Google Pixel 8 Pro (front camera).
    The microscope housings function as a spacer and ensure a fixed focal distance when pressed against the food surface.
    \item $100\times$ nominal magnification: The $100\times$ microscope by Apexel \cite{MS001100XLEDa} was used together with an iPhone SE 2nd Generation (main rear camera). Focusing is achieved manually by positioning the lens within the narrow depth of field ($<5$mm) relative to the sample surface.
\end{enumerate}

Notably, the utilized microscopes are highly affordable, with retail prices of approximately \$12 for the Jiusion \cite{Jiusion30XZoom} and 20€ ($\sim$\$24) for both the Apexel \cite{MS001100XLEDa} and Phonescope \cite{led-grower.euPhonescope30xZoom} models. We observed that visually identical $30\times$ microscope attachments are available in bulk for as little as \$1 \cite{30xLedlichtLupe}, suggesting that comparable image quality can be achieved with highly affordable and scalable hardware.

\begin{figure}[!t]
    \centering
    \begin{subfigure}[b]{0.495\linewidth}
        \centering
        \includegraphics[width=\linewidth]{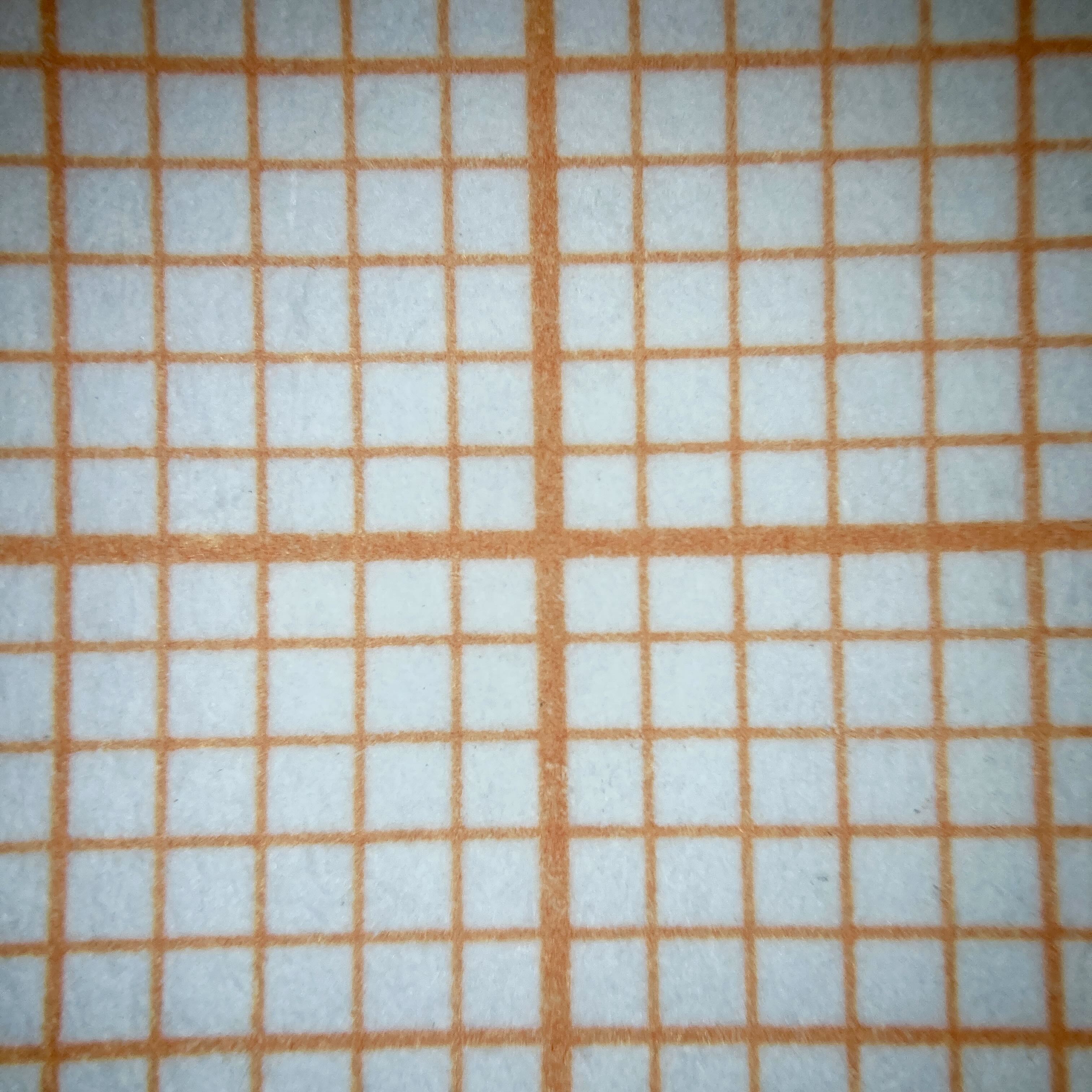}
        \caption{Apexel $100\times$ microscope.}
        \label{fig:fov_apexel}
    \end{subfigure}
    \hfill
    \begin{subfigure}[b]{0.495\linewidth}
        \centering
        \includegraphics[width=\linewidth]{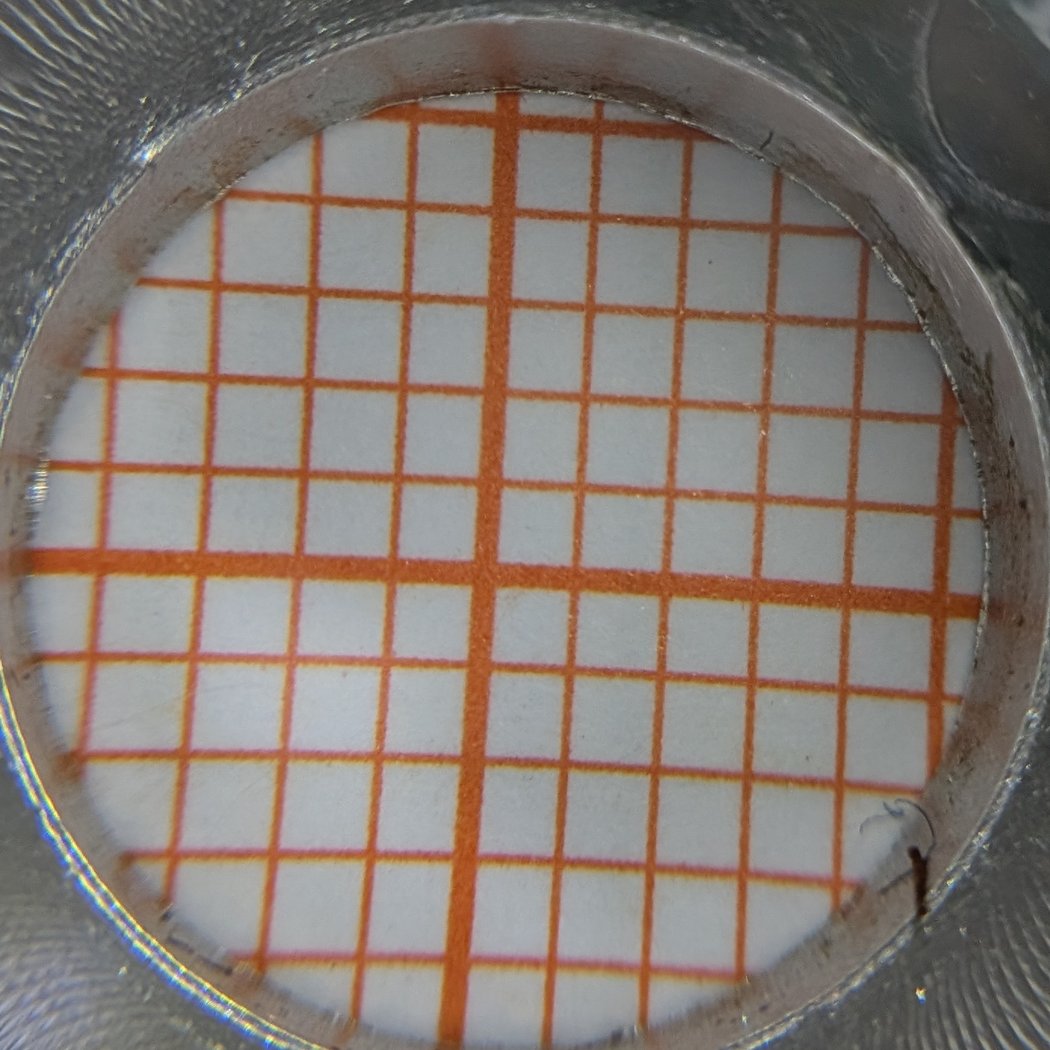}
        \caption{Phonscope $30\times$ microscope.}
        \label{fig:fov_yiwu}
    \end{subfigure}
    \caption{Effective field of view comparison using millimeter paper. Both images are cropped according to the preprocessing pipeline and show an effective FOV of $11\text{mm}\times 11\text{mm}$.}
    \label{fig:fov_calibration}
\end{figure}

Although nominal magnification differs, the effective field of view (FOV) after cropping is comparable across devices. This is because the vignette caused by the housing of the $30\times$ model covers $\sim80\%$ of the uncropped image. Figure~\ref{fig:fov_calibration} shows a photo of millimeter grid paper taken with the microscope attachments with the same cropping applied as in the preprocessing of our following experiments. The resulting effective FOV is $11 \text{mm}\times 11\text{mm}$ for all microscopes.

\subsubsection{Illumination}
Apart from environmental light sources, illumination was only provided by the microscopes' integrated LED lights: we employed a reflected light configuration (epi-illumination). This contrasts with the transmitted light (backlighting) approach used in \cite{treepongEarlyBreadMold2023b} for bread mold. Reflected light allows the imaging of opaque food matrices (e.g., cheese, carrots) without sample sectioning and aligns with typical usage conditions.

\subsubsection{Configuration and Environment}
\begin{figure}[!t]
    \centering
    \includegraphics[width=0.6\linewidth]{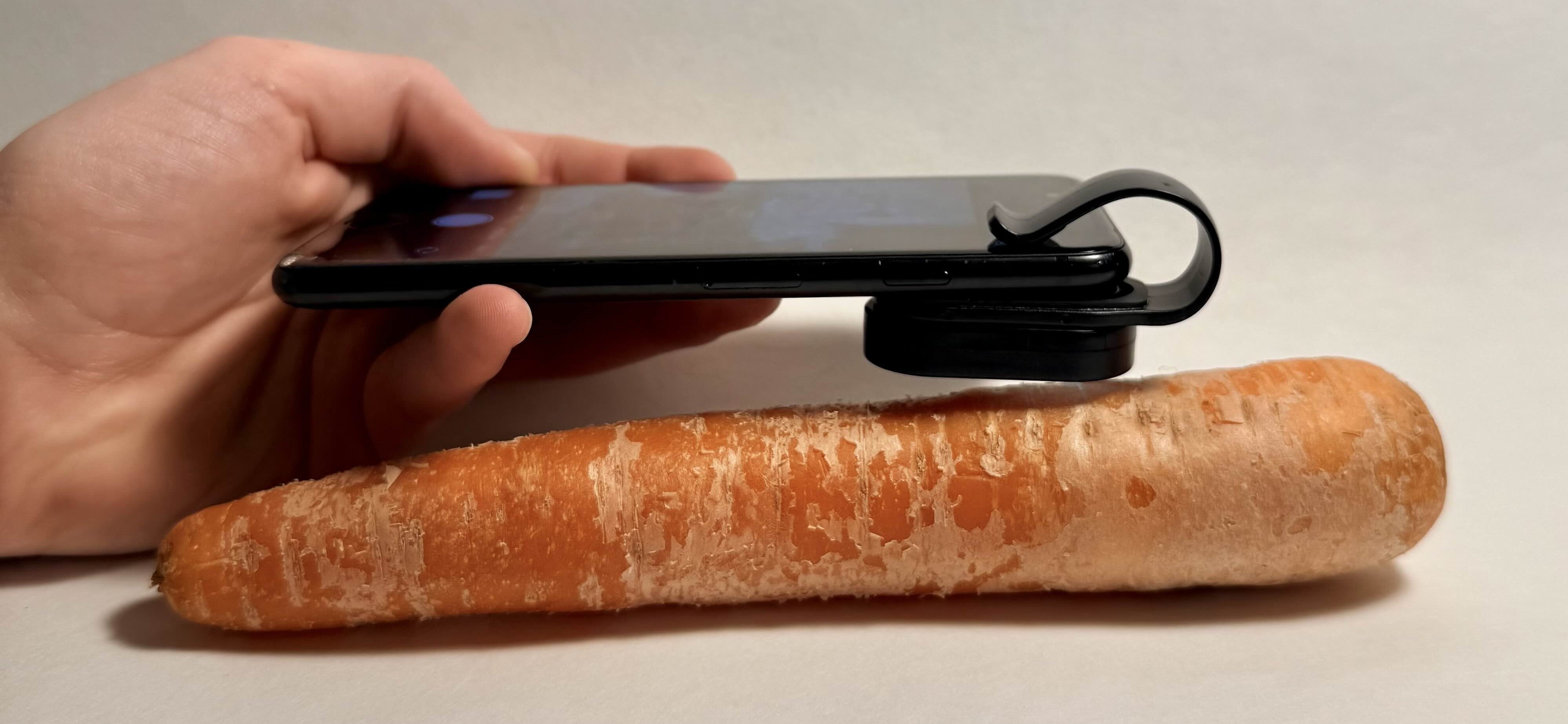} 
    \caption{Typical data acquisition setup mimicking consumer use. A smartphone equipped with the Apexel $100\times$ attachment is held manually over a carrot sample.}
    \label{fig:setup_carrot}
\end{figure}

To simulate realistic consumer behavior, all data collection was performed handheld, without the use of tripods or fixed mounts. Both the smartphone and food samples were manually rotated and positioned to acquire varied perspectives, as demonstrated in Figure~\ref{fig:setup_carrot}.

Images were captured as JPEGs using the native camera apps without digital zoom under varying ambient light conditions, notably daylight or artificial LED light. However, the internal microscope light stabilizes the lighting conditions.
Imaging conditions were kept as similar as possible for fresh and contaminated samples.

\FloatBarrier

\FloatBarrier

\subsection{Data Collection}

\subsubsection{Food Selection and Incubation}
Three food categories frequently susceptible to mold in domestic environments were identified: (a) bakery products, (b) fruits and vegetables, and (c) dairy products. To ensure the dataset’s practical utility, eleven distinct food items across these categories were selected: white toast, mixed grain bread, raspberries, blueberries, blackberries, cheese, tomatoes, onions, oranges, carrots, and cream cheese.

To guarantee realistic mold development and ensure successful contamination, cultivation protocols were specifically adapted for each food type. While most samples were stored in airtight plastic bags to maintain high humidity, incubation temperatures were varied between refrigeration for berries and room temperature for bread, vegetables, and dairy products. Unless otherwise stated, mold growth occurred spontaneously (natural growth). 

For all food categories except onions, a longitudinal capture approach was employed: baseline images (non-moldy) were initially recorded, followed by a second session where the same samples were photographed after a designated incubation period. This period ranged from 24 hours for some inoculated samples to 14 days for naturally decaying specimens.

A primary concern during the data collection process was the mitigation of spurious correlations. Since the 'mold' and 'no-mold' classes were recorded on different days, there was a significant risk that environmental or hardware-related variations could be misinterpreted by a model trained on the data as class-discriminatory features. For example, there might be dust particles appearing on the microscope lens during the second session.

To prevent such artifacts, rigorous maintenance protocols were implemented: all microscope lenses were thoroughly cleaned before and during each session, and identical smartphone-microscope pairings were strictly maintained across both recording dates. Furthermore, food-specific precautions were taken to address unique visual challenges, as detailed in the following subsections.

\subsubsection{Bakery Products}
The experiments were conducted with samples of white toast (Golden Toast "American Sandwich" and "Butter Toast") and mixed grain bread from a German bakery. Two distinct data collection protocols were applied to this category to ensure robustness against different stages of decay and user behaviors: \begin{itemize} \item \textbf{Systematic Inoculation Protocol:} A subset of bread samples (fresh toast and dried mixed bread) was manually inoculated with green mold spores harvested from pre-infected orange peels to simulate cross-contamination from other household items. These samples were moistened and stored overnight. Images were captured systematically: 10 images of the slice center, 5 of the edge, and 5 of the crust per slice. \item \textbf{Opportunistic Natural Protocol:} Additional toast samples were allowed to mold naturally at room temperature over several days. For these, we employed an "opportunistic" acquisition strategy, capturing overlapping images of large mold colonies from various angles. \end{itemize}

\subsubsection{Fruits and Vegetables}
This category encompasses berries (raspberries, blueberries, blackberries), tomatoes, onions, carrots, and oranges. \begin{itemize} \item \textbf{Berries:} Samples were stored in a refrigerator and allowed to develop white mold naturally over one week. To avoid lighting artifacts on the translucent skins, images were captured exclusively under natural daylight. We captured 7 images per individual berry across 6 distinct berries per type to avoid redundancy. \item \textbf{Vegetables and Oranges:} These items were stored at room temperature for 1-2 weeks in sealed bags. Data collection focused on structural diversity. Specific surface features (the stem and calyx region of tomatoes, soil residues on carrots, and corking on oranges) were systematically balanced within the dataset. A predefined ratio was maintained between images of typical surface textures and these specific features: a 3:1 distribution was utilized for tomatoes, while a 1:1 ratio was strictly followed for both carrots and oranges. Collection was performed opportunistically, scanning the surface for clear mold patches visible on the smartphone preview. \item\textbf{Onions:}
The onions were obtained at a supermarket, some of them already in a moldy condition. For data collection, an opportunistic protocol was followed.
\end{itemize}

\subsubsection{Dairy Products}
We processed Gouda cheese and cream cheese. \begin{itemize} \item \textbf{Hard Cheese (Gouda):} Samples were left to develop mold naturally, and images of the surface were captured.
\item \textbf{Cream Cheese:} To simulate cross-contamination, fresh cream cheese was smeared onto cardboard and inoculated with spores from infected fruit. After letting the mold grow for one week, we captured images across various textures, distinguishing between smooth, spread surfaces and crumbly, dried sections to prevent the model from overfitting to specific surface textures. \end{itemize}

\subsubsection{Image Acquisition and Post-Processing}
Across all categories, images were captured using the standard camera applications of the respective smartphones at $1\times$ zoom. Images were only saved if moldy structures were visible in the live preview.

No digital editing was applied during capture. Post-processing was limited to cropping the images to remove the circular vignette caused by the microscope lenses and resizing to $224\times224$ \si{px}.
MobileMold contains all sample images with and without post-processing.

\subsection{Dataset Statistics and Description}

\begin{figure}[!t]
        \centering

        \includegraphics[width=\linewidth]{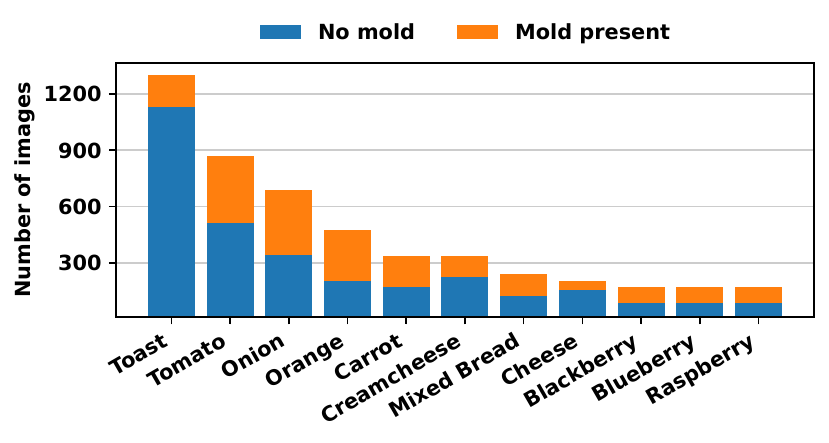}
    \caption{MobileMold per-food distribution of \textit{mold} vs.\ \textit{no mold}.}
    \label{fig:dataset_stats}
\end{figure}

MobileMold comprises 4{,}941 annotated smartphone microscopy images from 11 food types with a binary \textit{mold} label.
In total, 1{,}842 images contain visible mold (37.3\%) and 3{,}099 are labeled \textit{no mold} (62.7\%).
Figure~\ref{fig:dataset_stats} depicts the distribution of the food classes and mold.
2,681 images were captured using the Jiusion $30\times$ microscope paired with a Xiaomi Redmi Note 13, and 1,436 images using the Apexel $100\times$ paired with an iPhone SE (2nd gen.). The remaining 824 images were recorded with the Phonescope $30\times$, split evenly between a Samsung Galaxy S8+ (412) and a Google Pixel 8 Pro (412).

The released dataset follows a simple layout with a metadata table and a 1:1 mapping between original and preprocessed images: 

\vspace{3mm}
\dirtree{%
.1 MobileMold/.
.2 metadata.csv.
.2 train\_metadata.csv.
.2 val\_metadata.csv.
.2 test\_metadata.csv.
.2 original/ \DTcomment{4{,}941 images}.
.3 L10 - 48.jpeg.
.3 L10 - 25.jpeg.
.3 [...].
.2 cropped\_resized/ \DTcomment{4{,}941 images}.
.3 L10 - 48.jpeg.
.3 L10 - 25.jpeg.
.3 [...].
}
\vspace{3mm}

\begin{figure}[ht]
        \centering
        \includegraphics[width=0.8\linewidth]{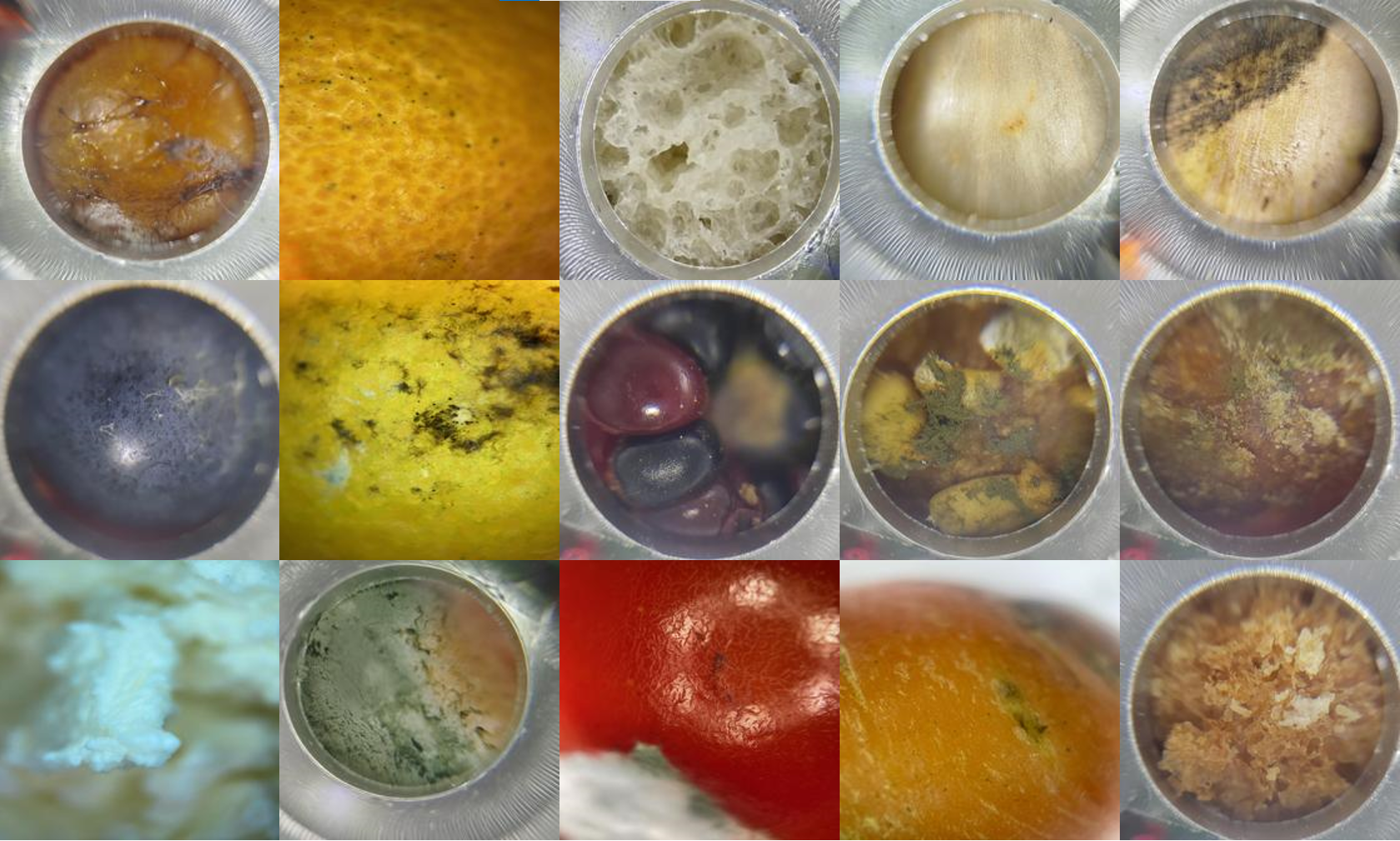}
    \caption{Samples of MobileMold.}
    \label{fig:header}
\end{figure}

We provide the dataset in both its raw (\textit{original}) and standardized (\textit{cropped\_resized}) formats. The latter follows the pre-processing described in Section \ref{sec:use-case}. The original filenames are preserved for the preprocessed samples in \textit{cropped\_resized}.
The main \texttt{metadata.csv} contains one row per image with the columns
\texttt{filename}, \texttt{mold} (True/False),
\texttt{food} (food class), \texttt{phone} (smartphone model), and
\texttt{microscope} (clip-on microscope model). 
To facilitate the reproducibility of our results, we provide the exact data split utilized in our later experiments as three CSV files (training, validation, and testing) with the same columns as the main metadata file. Fig. \ref{fig:header} illustrates some samples included in MobileMold.

\FloatBarrier

\FloatBarrier

\section{Dataset Use Case Application}
\label{sec:use-case}

The MobileMold dataset can support a diverse range of tasks and applications, including mold detection, food classification, mold localization, and segmentation. The images taken with clip-on lenses could also be utilized for texture analysis on a microscopic scale.  Furthermore, it serves as a valuable resource for downstream applications and transfer learning, such as starch morphology studies or the early-stage detection of bread mold as described in \cite{treepongEarlyBreadMold2023b}. To validate our dataset and demonstrate a practical application of MobileMold, this section presents the training and evaluation of image classifiers designed to detect mold and categorize food types. We report the performance of these models and utilize saliency maps to visualize whether the detected mold features are most relevant for the models' classification decisions. Finally, the mold detection models are integrated into a Flutter-based application, demonstrating the feasibility and usability of our approach for day-to-day consumer use on smartphones.

\subsection{Experiments} 

To automate the classification of mobile microscopy images, we evaluate different deep learning architectures. In our initial experiment, we benchmark the performance of seven pretrained models: DenseNet \cite{densenet}, MobileNet \cite{mobilenet}, ResNet \cite{He2015DeepRL}, VGG \cite{DBLP:journals/corr/SimonyanZ14a}, ShuffleNet \cite{Zhang2017ShuffleNetAE}, Swin Transformer \cite{swin}, and MaxViT \cite{maxvit}. The highest-performing architecture is selected as the backbone for the subsequent experiments. Concurrently, given our objective of enabling on-device mold detection, we also retain MobileNet for further testing due to its computational efficiency and optimization for mobile hardware.

The second experiment involves a comparison of data augmentation techniques to further enhance detection performance. We evaluate the prior best-performing model and MobileNet using several common augmentation strategies: Flip and Rotation, TrivialAugment \cite{Mller2021TrivialAugmentTY}, RandAugment \cite{randaugment}, AutoAugment \cite{autoaugment}, and AugMix \cite{hendrycks2020augmix}. For flipping and rotation, we apply horizontal and vertical flips with a 50\% probability and random 90-degree rotations. For the other methods, the default parameters were used. Data augmentation is only applied to the training data. The optimal augmentation strategy for each of the two models is then utilized in the final experiment.

Beyond binary mold detection, we also investigate the simultaneous identification of food class. To this end, we implement multi-task learning where mold detection and food classification are learned in parallel. Each pretrained model is modified to have two task-specific classification heads that follow the other unchanged layers as a set of shared feature extraction. This architecture allows the system to determine both the presence of mold and the food type in a single inference.

Furthermore, we compute saliency maps using vanilla gradients \cite{Simonyan14a}. To reduce visual noise, we apply SmoothGrad \cite{DBLP:journals/corr/SmilkovTKVW17} via Captum’s NoiseTunnel by adding Gaussian noise to the input and averaging the resulting attribution maps over 50 samples.

\subsection{Experimental Setup}

During preprocessing, we identify regions of interest within the raw images and perform cropping to eliminate artifacts such as the microscope vignette. We define specific pixel boundaries for the left, right, top, and bottom of each microscope model used to ensure precise cropping. Following this, all images are resized to a uniform resolution of $224\times224$ pixels.

To address the class imbalance introduced by an abundance of healthy toast samples, we randomly undersampled the toast class without mold to 513 instances. The dataset is partitioned into training, validation, and test sets using an 8:1:1 ratio. To ensure a balanced and rigorous evaluation, we maintain an equal distribution of samples with and without mold for each food class within the test set.

We assess classifier performance using six common classification metrics: accuracy (ACC), F1 score, Area Under the Curve (AUC), Precision-Recall AUC (PR AUC), Cohen’s kappa, and the Matthews Correlation Coefficient (MCC). While we report all metrics, we prioritize MCC for model selection and checkpoint evaluation. MCC has been shown to provide a more informative and reliable assessment of binary classification than accuracy or F1 score \cite{Chicco2020-kn}, which is particularly relevant for the task of mold detection. Furthermore, MCC aligns more closely with model performance on imbalanced datasets compared to PR AUC \cite{Imani2026}. Unlike ROC AUC, a high MCC value is only achieved when a classifier performs well across all four quadrants of the confusion matrix (sensitivity, specificity, precision, and negative predictive value) \cite{Chicco2023-rm}. Consequently, we follow recent recommendations to use MCC as the primary metric for classification \cite{Chicco2020-kn, Chicco2023-rm, Imani2026}, while reporting the others as supplementary indicators.

The hyperparameters and experimental settings are held constant across all experiments. We employ the Adam optimizer with a batch size of 64 and a learning rate of $\num{1e-5}$. We utilize cross-entropy loss while for multi-task learning, the total loss is calculated as the sum of the cross-entropy losses from both tasks. Models are trained for 60 epochs, and we select the checkpoint that achieves the highest MCC on the validation set for final evaluation on the test set. We used the PyTorch implementations \cite{Paszke2019pytorch} of the models (pretrained on ImageNet1k \cite{imagenet}) and data augmentation techniques, and the Captum library \cite{kokhlikyan2020captum} for generating saliency maps.

The code for dataset split generation, preprocessing, and all experiments is available at \url{https://github.com/MobileMold/mold-detection-baseline/}.

\subsection{Results and Discussion} 

In this section, we report the results of our experiments.

\begin{table}[]
\centering
\footnotesize
\begin{tabular}{l|llllll}
Model      & ACC    & F1     & AUC    & PR AUC & Kappa  & MCC    \\ \hline
DenseNet   & 0.9859 & 0.9858 & \textbf{0.9995} & \textbf{0.9995} & 0.9718 & 0.9720 \\
MobileNet  & 0.9836 & 0.9838 & 0.9992 & 0.9992 & 0.9671 & 0.9674 \\
Resnet     & 0.9648 & 0.9650 & 0.9965 & 0.9965 & 0.9296 & 0.9297 \\
VGG        & 0.9859 & 0.9860 & 0.9993 & 0.9993 & 0.9718 & 0.9719 \\
ShuffleNet & 0.9742 & 0.9740 & 0.9933 & 0.9913 & 0.9484 & 0.9485 \\
Swin       & \textbf{0.9883} & \textbf{0.9884} & 0.9989 & 0.9984 & \textbf{0.9765} & \textbf{0.9768} \\
MaxViT     & 0.9836 & 0.9837 & 0.9992 & 0.9992 & 0.9671 & 0.9672 \\ \hline
\end{tabular}
\caption{Model comparison on the MobileMold test set.}
\label{tab:model_comp}
\end{table}

\begin{table}[]
\centering
\footnotesize
\begin{tabular}{l|llllll}
Model      & ACC    & F1     & AUC    & PR AUC & Kappa  & MCC    \\ \hline
None            & 0.9883 & 0.9884 & 0.9989 & 0.9984 & 0.9765 & 0.9768 \\
Flip + Rotation & \textbf{0.9953} & \textbf{0.9953} & \textbf{0.9999} & \textbf{0.9999} & \textbf{0.9906} & \textbf{0.9907} \\
TrivialAugment  & \textbf{0.9953} & \textbf{0.9953} & 0.9996 & 0.9996 & \textbf{0.9906} & 0.9906 \\
RandAugment     & 0.9930 & 0.9929 & 0.9998 & 0.9998 & 0.9859 & 0.9859 \\
AutoAugment     & 0.9906 & 0.9907 & 0.9995 & 0.9995 & 0.9812 & 0.9813 \\
AugMix          & 0.9930 & 0.9929 & 0.9998 & 0.9998 & 0.9859 & 0.9859 \\ \hline
\end{tabular}
\caption{Augmentation comparison on the MobileMold test set for Swin Transformer.}
\label{tab:aug_swin}
\end{table}

\begin{table}[]
\centering
\footnotesize
\begin{tabular}{l|llllll}
Model      & ACC    & F1     & AUC    & PR AUC & Kappa  & MCC    \\ \hline
None            & 0.9836 & 0.9838 & 0.9993 & 0.9993 & 0.9672 & 0.9675 \\
Flip + Rotation & 0.9695 & 0.9695 & 0.9978 & 0.9979 & 0.9390 & 0.9390 \\
TrivialAugment  & 0.9907 & 0.9907 & \textbf{0.9999} & \textbf{0.9999} & 0.9813 & 0.9813 \\
RandAugment     & 0.9907 & 0.9907 & 0.9995 & 0.9995 & 0.9813 & 0.9813 \\
AutoAugment     & 0.9883 & 0.9883 & 0.9996 & 0.9996 & 0.9766 & 0.9766 \\
AugMix          & \textbf{0.9954} & \textbf{0.9954} & \textbf{0.9999} & \textbf{0.9999} & \textbf{0.9907} & \textbf{0.9907} \\ \hline
\end{tabular}
\caption{Augmentation comparison on the MobileMold test set for MobileNet.}
\label{tab:aug_mobilenet}
\end{table}

\begin{table}[]
\centering
\footnotesize
\setlength{\tabcolsep}{5pt}
\begin{tabular}{l l|llllll}
Model & Task & ACC & F1 (macro) & AUC (OvR) & Kappa & MCC \\ \hline

\multirow{2}{*}{Swin}
& Mold & 0.9883 & 0.9883 & 0.9998 & 0.9765 & 0.9766 \\
& Food & 1.0000 & 1.0000 & 1.0000 & 1.0000 & 1.0000 \\ \hline

\multirow{2}{*}{MobileNet}
& Mold & 0.9906 & 0.9906 & 0.9997 & 0.9812 & 0.9812 \\
& Food & 0.9977 & 0.9964 & 1.0000 & 0.9973 & 0.9973 \\ \hline

\end{tabular}
\caption{Test performance for multitask learning on mold detection (binary) and food type classification (multiclass). F1 is macro-averaged, multiclass AUC is reported using One-versus-rest.}
\label{tab:multitask}
\end{table}

\paragraph{Architecture Comparison}
Table \ref{tab:model_comp} summarizes the results for binary mold detection across the seven models. The Swin Transformer emerged as the top-performing model, achieving the highest MCC (0.9768). However, it is noteworthy that even lightweight models optimized for mobile deployment, such as MobileNet and ShuffleNet, achieved accuracies exceeding $97\%$. The consistently high AUC and PR AUC values (all > 0.99) across nearly all architectures suggest that the features captured in the microscopic images are highly discriminative. 

\paragraph{Impact of Data Augmentation}
We tested various data augmentation strategies to further improve the performance of the top-performing (Swin Transformer) and mobile-friendly model (MobileNet).

For the Swin Transformer (Table \ref{tab:aug_swin}), basic geometric transformations (Flip + Rotation) provided a significant performance boost, reaching an MCC of 0.9907. Orientation invariance seems to be a primary factor in generalization. 

A quite different result is obtained for MobileNet (Table \ref{tab:aug_mobilenet}): Flip + Rotation even led to a slightly degraded performance while AugMix significantly enhanced the model's performance, achieving the same MCC as Swin Transformer. This may indicate that for smaller CNN-based architectures, more sophisticated mixing and diverse augmentation chains are helpful to train a model that generalizes well.

\paragraph{Multitask Learning}
Finally, we investigated how adding the task of food classification impacted the performance on mold detection (see Table \ref{tab:multitask}). Both models achieve a much higher performance on the food classification task than on mold detection, with the Swin Transformer scoring perfect accuracy. This might be due to the food being easier to visually distinguish than to recognize mold. On the mold task, the MCC decreased from 0.9907 to 0.9766. 
The MobileNet shows similar behavior, achieving near-perfect MCC (0.9907) on food classification, but reduced performance on the mold task (0.9812). One potential explanation is that the high baseline accuracy in mold detection left little room for improvement, meaning the addition of a secondary task likely induced inter-task interference across shared representations. Despite this potential for interference, the performance metrics under the multi-task learning framework remain exceptionally high. This additional granularity provides significant value for end-users. For instance, identifying the specific variety of bread or cheese may have distinct safety or dietary implications in the event of contamination \cite{Gomez2023-ea,Altafini2021-je,Oueslati2020-qb,Pavicich2024-pz}.

\begin{figure}[!t]
    \centering
    \includegraphics[width=0.65\linewidth, angle=-90]{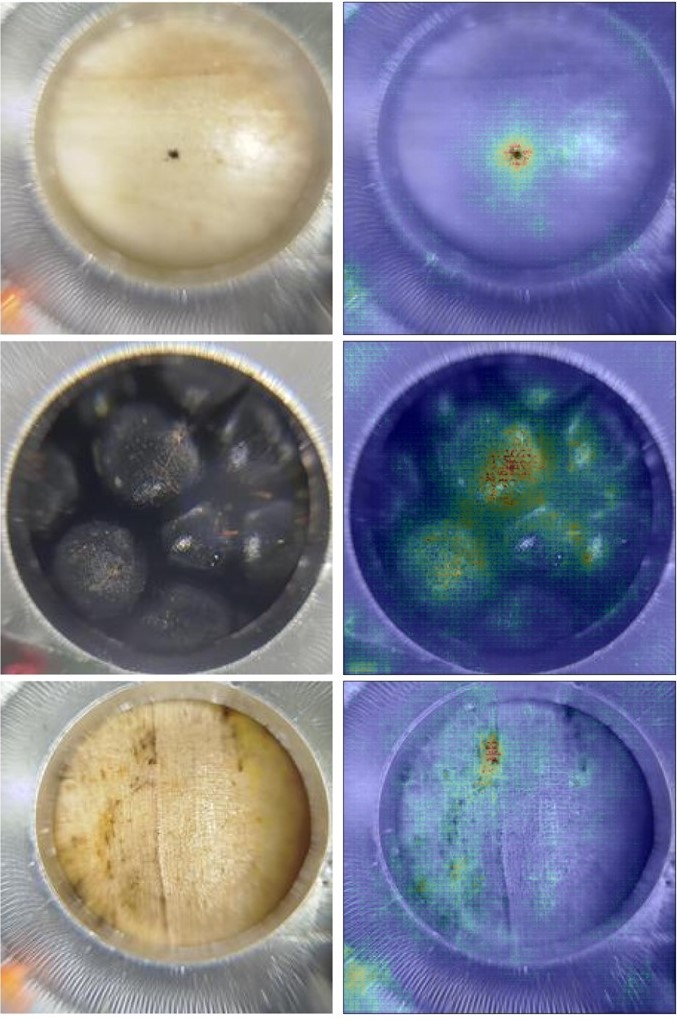} 
    \caption{Saliency maps of MobileNet for 3 random samples of the test set with mold.}
    \label{fig:saliency}
\end{figure}

We generated saliency maps for test-set samples. Fig. \ref{fig:saliency} shows MobileNet saliency maps for three randomly selected mold-positive images that were correctly classified. In these examples, the highlighted regions largely overlap with visually apparent mold areas, suggesting that the model’s prediction is influenced by mold-related image regions.
While this provides qualitative evidence consistent with the intended behavior, saliency maps are not a definitive proof of causal feature usage and do not fully rule out reliance on background artifacts or other spurious correlations.

\begin{figure}[!htbp]
    \centering
    \includegraphics[width=0.45\linewidth]{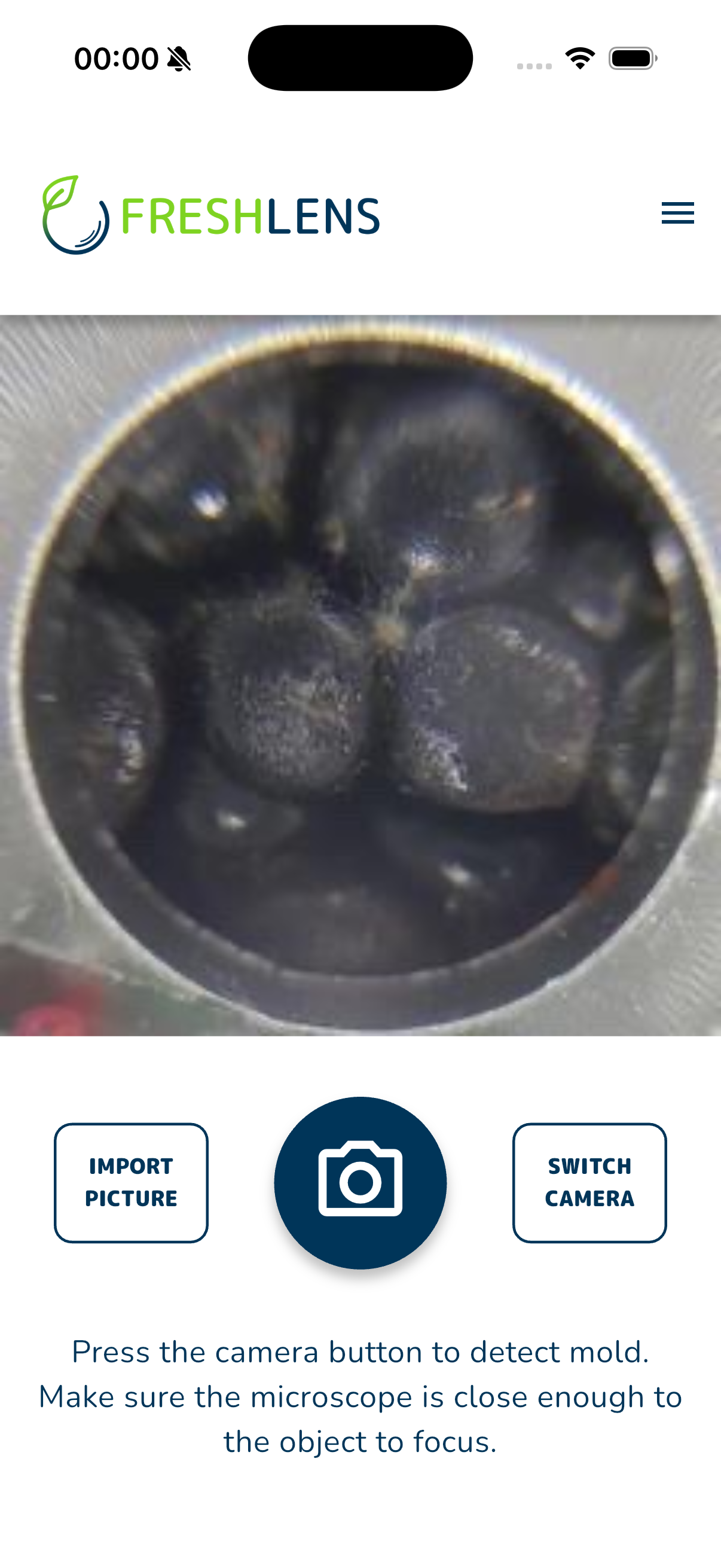}
    \qquad
    \includegraphics[width=0.45\linewidth]{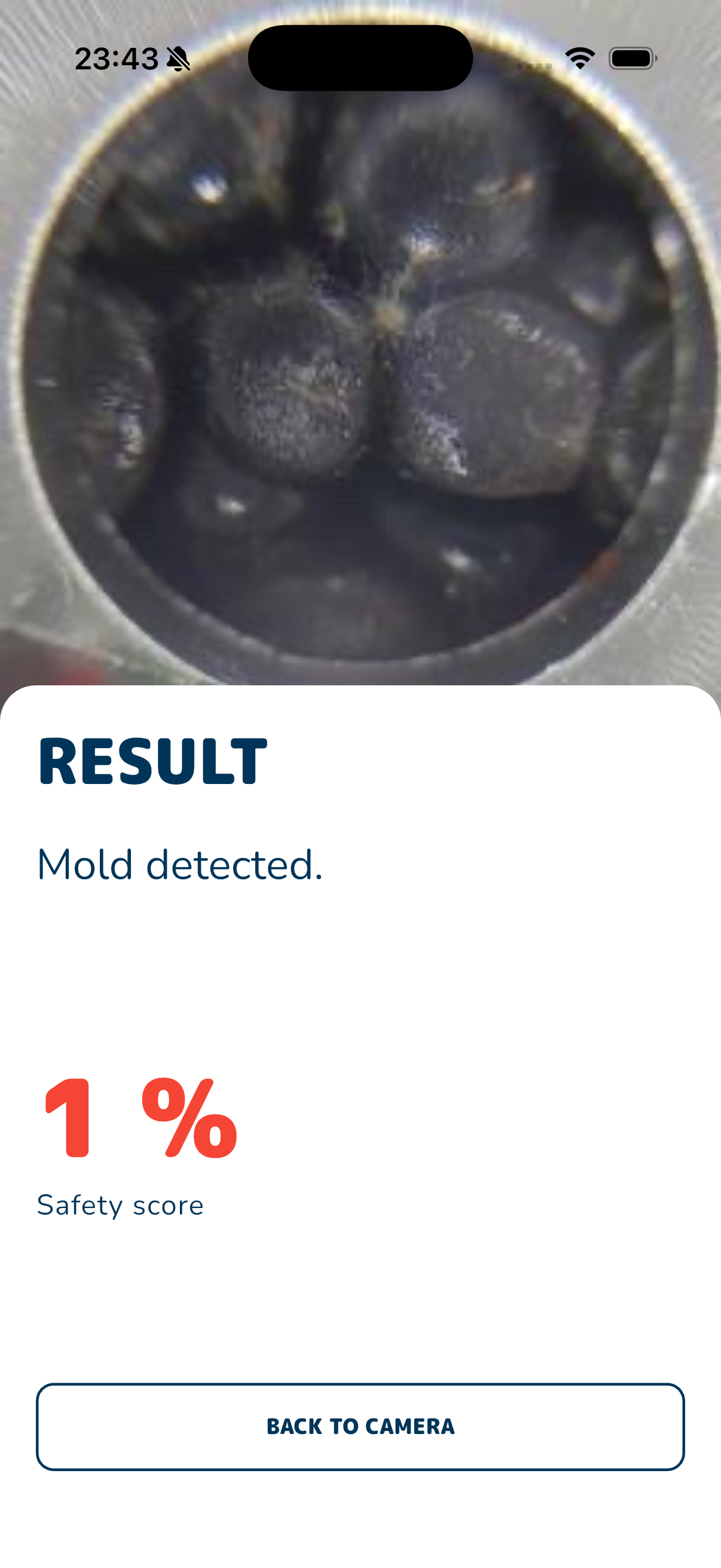}
    \caption{Main screen interface (left) and analysis screen showing mold detection results (right) of our app.}
    \label{fig:screens}
\end{figure}

Lastly, we developed a Flutter app that consumers with a microscope attachment can use to take a picture and let the fine-tuned MobileNet analyze it, outputting the softmax probability. The interface can be seen in Fig. \ref{fig:screens} and the code is available at \url{https://github.com/MobileMold/freshlens-app}.

\FloatBarrier

\section{Conclusion}
In this paper, we introduced MobileMold, a novel dataset bridging the gap between professional microscopy and mobile sensing. By utilizing low-cost attachments, we demonstrate that smartphones can function as powerful, non-destructive tools for rapid food safety assessment. Our work overcomes the constraints of conventional methods, providing a democratized solution for consumer-grade microscopic sensing in everyday environments. Our benchmarking of deep learning models establishes robust baselines for mold detection and food classification. The inclusion of saliency maps provides post-hoc explanations, suggesting that our classifiers prioritize relevant fungal spots over environmental artifacts. Furthermore, the integration of these models into a Flutter-based application demonstrates the potential real-world viability of MobileMold. By releasing MobileMold as an open-source resource, we provide a unique asset for the multimedia community to explore smartphone attachments, micro-scale texture analysis, domain adaptation, and on-device inference. Future work will focus on expanding the dataset’s diversity and investigating the temporal dynamics of mold growth. Ultimately, MobileMold contributes to the development of accessible, mobile-first technologies that enhance food security and empower consumers globally.



\bibliographystyle{ACM-Reference-Format}
\bibliography{refs_mobilemold.bib}

\appendix

\end{document}